\documentclass[10pt,twocolumn,letterpaper]{article}
\usepackage{amsmath}
\usepackage{amssymb}
\usepackage{booktabs}
\usepackage{enumerate}
\usepackage{epsfig}
\usepackage{graphicx}
\usepackage{multirow}
\usepackage{times}
\usepackage{wacv}
\usepackage{adjustbox}
\usepackage{subfig}

\graphicspath{{figures/}}                                                                                                                                                                                   
\def\wacvPaperID{1434} % Enter the WACV Paper ID here

\wacvalgorithmstrack   % Uncomment this line if you are submitting to the Algorithms Track.
\wacvfinalcopy % *** Uncomment this line for the final submission

\ifwacvfinal
\usepackage[breaklinks=true,bookmarks=false,hidelinks]{hyperref}
\else
\usepackage[pagebackref=true,breaklinks=true,colorlinks,bookmarks=false]{hyperref}
\fi

\begin{document}

\title{Single Image Super-Resolution via a Dual Interactive Implicit Neural Network}

\author{Quan H. Nguyen and William J. Beksi\\
The University of Texas at Arlington\\
Arlington, TX, USA\\
{\tt\small quan.nguyen4@mavs.uta.edu, william.beksi@uta.edu}
}

\maketitle
\thispagestyle{empty}

%%%%%%%%%%%%%%%%%%%%%%%%%%%%%%%%%%%%%%%%%%%%%%%%%%%%%%%%%%%%%%%%%%%%%%%%%%%%%%%%
\begin{abstract}
In this paper, we introduce a novel implicit neural network for the task of
single image super-resolution at arbitrary scale factors. To do this, we
represent an image as a decoding function that maps locations in the image
along with their associated features to their reciprocal pixel attributes.
Since the pixel locations are continuous in this representation, our method can
refer to any location in an image of varying resolution. To retrieve an image
of a particular resolution, we apply a decoding function to a grid of locations
each of which refers to the center of a pixel in the output image.  In contrast
to other techniques, our dual interactive neural network decouples content and
positional features. As a result, we obtain a fully implicit representation of
the image that solves the super-resolution problem at (real-valued) elective
scales using a single model. We demonstrate the efficacy and flexibility of our
approach against the state of the art on publicly available benchmark datasets. 
\end{abstract}

%%%%%%%%%%%%%%%%%%%%%%%%%%%%%%%%%%%%%%%%%%%%%%%%%%%%%%%%%%%%%%%%%%%%%%%%%%%%%%%%
\section{Introduction} 
\label{sec:introduction}
Single image super-resolution (SISR) is a fundamental low-level computer vision
problem that aims to recover a high-resolution (HR) image from its
low-resolution (LR) counterpart. There are two main reasons for performing
SISR: (i) to enhance the visual quality of an image for human consumption, and
(ii) to improve the representation of an image for machine perception. SISR has
many practical applications including robotics, remote sensing, satellite
imaging, thermal imaging, medical imaging, and much more
\cite{yue2016image,yang2017image}. Despite being a challenging and ill-posed
subject, SISR has remained a crucial area of study in the research community. 

Recent deep learning approaches have provided high-quality SISR results
\cite{bai2020survey,wang2020brief}. In perception systems, images are
represented as 2D arrays of pixels whose quality, sharpness, and memory
footprint are controlled by the resolution of the image. Consequently, the scale
of the generated HR image is fixed depending on the training data. For example,
if a neural network is trained to recover HR images of $\times 2$ scale then it
only performs well on what it is trained for (\ie, performance will be poor on
$\times 3$, $\times 4$, or other scales). Thus, instead of training multiple
models for various resolutions, it can be extremely useful in terms of
practicality to have a \textit{single SISR architecture} that handles
\textit{arbitrary scale factors}.  This is especially true for embedded vision
platforms (e.g., unmanned ground/aerial vehicles) with multiple on-board cameras
that must execute difficult tasks using limited computational resources. 

\begin{figure}
\centering
\includegraphics[width=.9\linewidth]{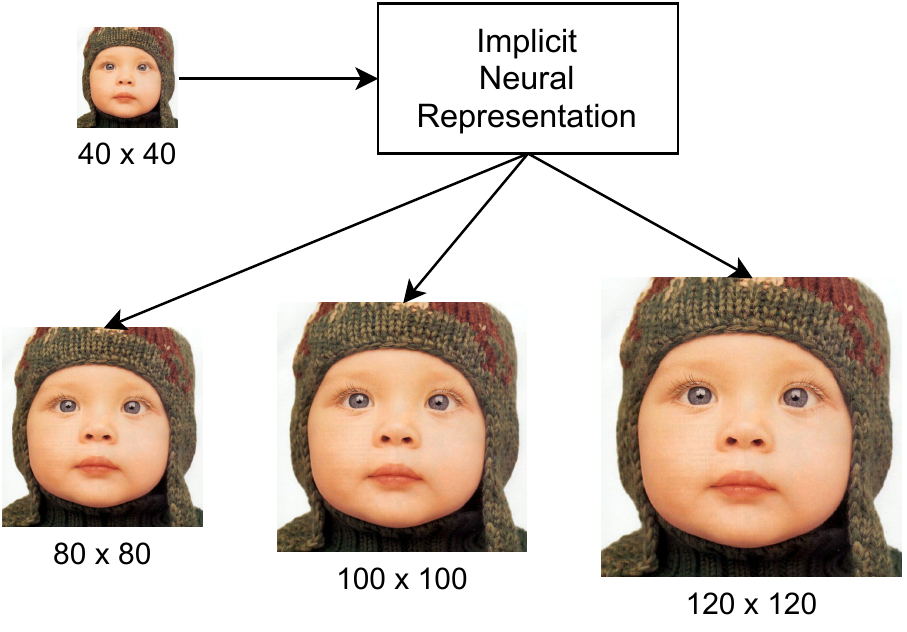}
\caption{Our proposed dual interactive implicit neural network (DIINN) is
capable of producing images of \textit{arbitrary resolution}, using a
\textit{single trained model}, by capturing the underlying implicit
representation of the input image.}
\label{fig:concept_overview}
\end{figure}

The notion of an implicit neural representation, also known as coordinate-based
representation, is an active field of research that has yielded substantial
results in modeling 3D shapes \cite{atzmon2020sal,chabra2020deep,jiang2020local,
peng2020convolutional,ramon2021h3d}. Inspired by these successes, learning
implicit neural representations of 2D images is a natural solution to the SISR
problem since an implicit system can produce output at arbitrary resolutions.
While this idea has been touched upon in several works
\cite{bemana2020x,sitzmann2020implicit,chen2021learning,mehta2021modulated}, in
this paper we propose a more expressive neural network for SISR with
\textit{significant improvements} over the existing state of the art,
Figure~\ref{fig:concept_overview}. Our contributions are summarized as follows.
\begin{itemize}
  \item We develop a novel dual interactive implicit neural network (DIINN) for
  SISR that handles image content features in a modulation branch and positional
  features in a synthesis branch, while allowing for interactions between
  the two.
  \item We learn an implicit neural network with a pixel-level representation,
  which allows for locally continuous super-resolution synthesis with respect to
  the nearest LR pixel.
  \item We demonstrate the effectiveness of our proposed network by setting new
  benchmarks on public datasets.
\end{itemize}

Our source code is available at \cite{diinn2023}. The remainder of this paper is
organized as follows. Related research is discussed in
Section~\ref{sec:related_work}. In
Section~\ref{sec:dual_interactive_implicit_neural_network}, we present a
detailed description of our model for SISR at arbitrary scales using an implicit
representation. Experimental results are presented in
Section~\ref{sec:experiments}. The paper concludes in
Section~\ref{sec:conclusion} and discusses future work.

\begin{figure*}
\centering
\includegraphics[width=.95\textwidth]{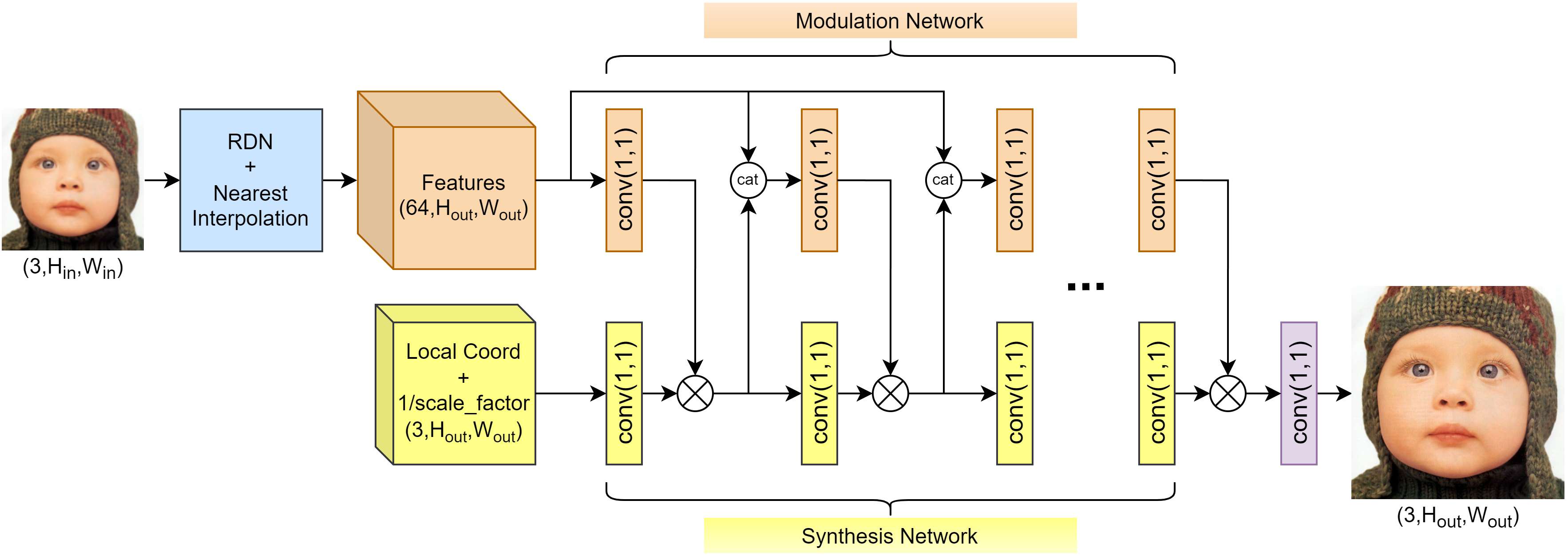}
\caption{An overview of our DIINN architecture. Note that each layer in the
modulation network is immediately followed by a ReLU activation, \eqref{eq:mpn1}
and \eqref{eq:mpn3}. Similarly, each layer in the synthesis network is followed
by a sine activation, \eqref{eq:mpn2} and \eqref{eq:mpn4}.}
\label{fig:overview}
\end{figure*}

%%%%%%%%%%%%%%%%%%%%%%%%%%%%%%%%%%%%%%%%%%%%%%%%%%%%%%%%%%%%%%%%%%%%%%%%%%%%%%%%
\section{Related Work} 
\label{sec:related_work}
This section highlights pertinent literature on the task of SISR. First, we
discuss deep learning techniques for SISR. Then, we provide an overview of
implicit neural representations. Lastly, we cite the nascent domain of implicit
representations for images. The SISR problem is an ill-defined problem in the
sense that there are many possible HR images that can be downsampled to a single
LR image. In this work, we focus on learning deterministic mappings, rather than
stochastic mappings (\ie, generative models). In general, the input to an SISR
system is an LR image, and the output is a super-resolved (SR) image that may or 
may not have the same resolution as a target HR image.

\subsection{Deep Learning for SISR}
Existing work on SISR typically utilizes convolutional neural networks (CNNs)
coupled with upsampling operators to increase the resolution of the input image.

\subsubsection{Upscaling $+$ Refinement}
SRCNN \cite{dong2015image}, VDSR \cite{kim2016accurate}, and DRCN
\cite{kim2016deeply} first interpolate an LR image to a desired resolution using
bicubic interpolation, followed by a CNN-based neural network to enhance the
interpolated image and produce an SR image. The refining network acts as a
nonlinear mapping, which aims to improve the quality of the interpolation.
These methods can produce SR images at arbitrary scales, but the performance is
severely affected by the noise introduced during the interpolation process. The
refining CNNs also have to operate at the desired resolution, thus leading to
a longer runtime.

\subsubsection{Learning Features $+$ Upscaling}
Methods following this approach first feed an LR image through a CNN to obtain a
deep feature map at the same resolution. In this way, the CNNs are of cheaper
cost since they are applied at LR, which allows for deeper architectures. Next,
an upscaling operator is used to produce an SR image. The most common upscaling
operators are deconvolution (FSRCNN \cite{dong2016accelerating}, DBPN
\cite{haris2018deep}), and sub-pixel convolution (ESPCN \cite{shi2016real}, EDSR
\cite{lim2017enhanced}). It is also possible to perform many iterations of
\textit{learning features $+$ upscaling} and explicitly exploit the relationship
between intermediate representations \cite{haris2018deep}. These methods only
work with integer scale factors and produce fixed-sized outputs. 

EDSR \cite{lim2017enhanced} attempts to mitigate these problems by training a
separate upscaling head for each scale factor. On the other hand, Meta-SR
\cite{hu2019meta} is among the first attempts to solve SISR at arbitrary
real-valued scale factors via a soft version of the sub-pixel convolution. To
predict the signal at each pixel in the SR image, Meta-SR uses a meta-network to
determine the weights for features of a $(3\times3)$ window around the nearest
pixel in the LR image. Effectively, each channel of the predicted pixel in the
SR image is a weighted sum of a $(C\times3\times3)$ volume, where $C$ is the
number of channels in the deep feature map. While Meta-SR has a limited
generalization capability to scale factors larger than its training scales, it
can be viewed as a hybrid implicit/explicit model.
%which explicitly maps LR to HR features and implicitly conditions the mapping
%with relative offsets between LR and HR pixels via the meta-network.

\subsection{Implicit Neural Representations}
Implicit neural representations are an elegant way to parameterize signals
continuously in comparison to conventional representations, which are usually
discrete. Chen \etal \cite{chen2019learning}, Mescheder \etal
\cite{mescheder2019occupancy}, and Park \etal \cite{park2019deepsdf} are among
the first to show that implicit neural representations outperform 3D
representations (\eg, meshes, voxels, and point clouds) in 3D modeling. Many
works that achieve state-of-the-art results in 3D computer vision have followed.
For example, Chabra \etal \cite{chabra2020deep} learned local shape priors for
the reconstruction of 3D surfaces coupled with a deep signed distance function.
A new implicit representation for 3D shape learning called a \textit{neural
distance field} was proposed by Chibane \etal \cite{chibane2020ndf}. Jiang \etal
\cite{jiang2020local} leveraged voxel representations to enable implicit
functions to fit large 3D scenes, and Peng \etal \cite{peng2020convolutional}
increased the expressiveness of 3D scenes with various convolutional models. It
also is possible to condition the implicit neural representations on the input
signals \cite{chabra2020deep,chibane2020implicit,jiang2020local,
peng2020convolutional}, which can be considered as a hybrid implicit/explicit
model.

\subsubsection{Implicit Neural Representations of 2D Images}
Despite there being many uses of implicit neural representations in 3D computer
vision, to the extent of our knowledge, it has been under-explored in the domain
of 2D imaging. Parameterizing images implicitly via neural networks can be
traced back to Stanley \etal \cite{stanley2007compositional} in 2007.  Various
types of signals, including images, are implicitly represented with neural
networks using periodic activation functions by Sitzmann \etal
\cite{sitzmann2020implicit}. Bemana \etal \cite{bemana2020x} learned
representations of changes in view, time, and light using Jacobians of pixel
positions to naturally interpolate images. Tancik \etal
\cite{tancik2020fourfeat} showed that a coordinate-based multilayer perceptron
(MLP) can implicitly synthesize 2D images with sharpness by transforming the
coordinate inputs with a Fourier feature mapping. These works have limited
generalizability to different instances, especially as the complexity of the
data increases. Recently, Mehta \etal \cite{mehta2021modulated} proposed to
modulate a synthesis network with a separate network whose inputs are local
features, which enables generalization while maintaining high fidelity.

\subsubsection{Implicit Neural Representations for SISR}
Learning implicit neural representations of 2D images is immediately useful for
SISR since it can sample the pixel signals at any location in the spatial
domain. Chen \etal \cite{chen2021learning} proposed a local implicit image
representation in which 2D deep feature maps of the input images coupled with an
ensemble of local predictions ensure a smooth (\ie, continuous) transition
between different locations. Ma \etal \cite{ma2022recovering} extended
\cite{chen2021learning} by enforcing sharpness constraints and additionally
imposing multiple loss functions (including L1, perceptual
\cite{johnson2016perceptual}, and generative adversarial
\cite{goodfellow2014generative} losses) versus the L1 loss in
\cite{chen2021learning} to generate perceptually-pleasant details. In contrast
to \cite{chen2021learning}, we improve the implicit decoding function by
decoupling the content and the positional features as proposed by
\cite{mehta2021modulated} at pixel-level representations.

\begin{figure}
\centering
\includegraphics[width=.4\textwidth]{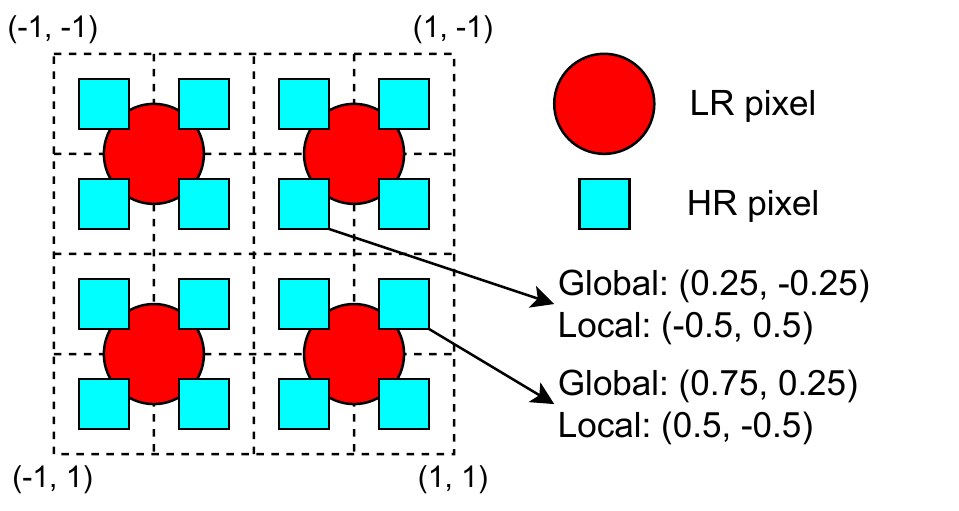}
\caption{An image at {\color{red} $(2,2)$} and {\color{cyan} $(4,4)$}
resolutions. We consider pixels as squares rather than points, the global and
local coordinates reference the centers of pixel squares with respect to the
center of the image and the center of the nearest LR pixel, respectively. Both
are scaled to $[-1,1]^2$.}
\label{fig:pixel_coordinates}
\end{figure}

%%%%%%%%%%%%%%%%%%%%%%%%%%%%%%%%%%%%%%%%%%%%%%%%%%%%%%%%%%%%%%%%%%%%%%%%%%%%%%%%
\section{Dual Interactive Implicit Neural Network} 
\label{sec:dual_interactive_implicit_neural_network}
In this section, we present the core components of our proposed model for SISR.
The network consists of the following two parts: an encoder and a dual
interactive implicit decoder, Figure~\ref{fig:overview}. The encoder learns the
content of the LR image and produces a deep feature map. The implicit decoder
predicts the signal (\ie, the $(r,g,b)$ values) at any query location within the
image space, conditioned on the associated features provided by the encoder. For
a target resolution, we query the signal of every pixel in the SR output image.
Our network can be expressed as
\begin{equation} \label{eq:framework}
  S_{(x,y)} = f_\theta(g_\gamma(L)_{(x,y)}, p),
\end{equation}
where $S_{(x,y)}$ is the SR predicted signal at $(x,y)$, $L$ is the LR image,
$g_\gamma(\cdot)$ is the encoder function parameterized by $\gamma$,
$g_\gamma(L)_{(x,y)}$ is the extracted features at $(x,y)$ (referred to as
content features), $p$ is the positional features, and $f_\theta(\cdot, \cdot)$
is the implicit decoder function parameterized by $\theta$. Note that our method
considers pixels of an image as squares rather than points and the location of a
pixel references its center.

\subsection{Encoder} 
\label{subsec:method-encoder}
The encoder supplies the decoder with a content feature representation
associated with each query location within the image. The encoder used in our
method could be a CNN, similar to those used in previous works
\cite{chen2021learning,hu2019meta}, without any upscaling module. The input is
an LR image and its output is a deep feature map that preserves the spatial
content of the image. With proper settings (\eg, kernel size, padding, \etc),
the output has the same spatial size as the input image. Thus, given a set of
coordinates for query locations, we can sample its corresponding features via
interpolation.

\subsubsection{Feature Unfolding} 
Following previous works \cite{chen2021learning,hu2019meta}, before
interpolation we enrich the feature map by applying feature unfolding with a
kernel size of $3$. That is, we concatenate the features of a $3\times3$
neighborhood around each pixel of the deep feature map, increasing the number of
channels by 9. Since our implicit decoder handles continuous queries and the
predictions are locally continuous, we choose nearest interpolation to avoid the
additional smoothness constraint as well as artifacts that come with bilinear or
bicubic interpolation. For any query location, the encoder effectively supplies
the decoder with the nearest features in the deep feature map. As expressed in
\eqref{eq:framework}, $g_\gamma(L)_{(x,y)}$ learns a feature map of $L$ and
allows for querying content features (via nearest interpolation) associated with
a coordinates $(x,y)$.

\subsection{Decoder} 
\label{subsec:method-decoder}
The decoder predicts the signal at every pixel in the SR image at a target
resolution. Our implicit decoder uses both the content features (from the
encoder) and the positional features for prediction.  For SISR, the positional
features of a query location typically contain information about its relative
position to the nearest LR pixel (center) and some information about the scale
factor \cite{hu2019meta}. We refer to the coordinates that reference the
position of a pixel with respect to the center of the image as the
\textit{global} coordinates. Conversely, we refer to the coordinates that
reference the position of a pixel with respect to its nearest LR pixel as the
\textit{local} coordinates. In our method, the \textit{global} coordinates
allow for uniquely identifying each pixel and computing its nearest LR pixel
while the \textit{local} ones are used as input to the decoder.
Figure~\ref{fig:pixel_coordinates} shows examples of \textit{global} and
\textit{local} coordinates.

\subsubsection{Modulated Periodic Activations Network} 
Recently, neural networks with periodic activation functions
\cite{sitzmann2020implicit,mehta2021modulated} exhibit excellent performance in
reconstructing high-fidelity images and videos. We use a similar dual MLP
architecture, also known as a modulated periodic activations neural network
\cite{mehta2021modulated}, for our decoder with adjustments to address the SISR
task. Our decoder of $N$ layers (\ie, $N$ layers of modulation and $N$ layers of
synthesis) can be written recursively as
\begin{align}
  m_0 &= ReLU(w_0 z + b_0), \label{eq:mpn1}\\ 
  s_0 &= m_0 \odot sin(w'_0 p + b'_0), \label{eq:mpn2}\\
  m_i &= ReLU(w_i [ s_{i-1} \; z ] + b_i), \label{eq:mpn3}\\
  s_i &= m_i \odot sin(w'_i s_{i-1} + b'_i), \label{eq:mpn4}
\end{align}
where $w$, $w'$, $b$, and $b'$ are the weights and biases, $z$ is the content
features, and $p$ is the positional features.  The last output of $s_i$ is then
passed through a final dense layer (without activation) to output the predicted
SR signal.

The two main differences between our method and \cite{mehta2021modulated} are
the following.
\begin{enumerate}[(i)]
  \item The $z$ vector in the image reconstruction experiment in
  \cite{mehta2021modulated} represents a patch while ours represents a pixel in
  the LR image. Typically, coarse-grain features are associated with higher
  levels of semantic information and thus may miss low-level details (\eg,
  edges, corners, \etc), which are \textit{crucial} for SISR. Additionally,
  pixel-level representations have consistently performed well in the SISR
  literature \cite{lim2017enhanced,hu2019meta,chen2021learning}. Therefore, we
  opt for the finest-grain representation (\ie, pixel-level). The architecture
  of our encoder reflects this choice.
  \item In \eqref{eq:mpn3}, we use a concatenation of the output of the previous
  synthesis layer (instead of the previous modulation layer in
  \cite{mehta2021modulated}) and the latent feature vector as inputs for the
  latter modulation layer. We argue that the outputs of the synthesis network
  are progressively more refined toward different query locations and therefore
  provide better information to the modulation network while the latent feature
  vector serves as residual feedback. We show the benefits of our modification
  over \cite{mehta2021modulated} in Section~\ref{sec:experiments}.
\end{enumerate}

\begin{table*}
\centering
\begin{adjustbox}{scale=.9}
    \begin{tabular}{c|c|c c c | c c c | c c c}
        \multirow{2}{4em}{Dataset} & Scale & \multicolumn{3}{c|}{$\times 2$} & \multicolumn{3}{c|}{$\times 3$} & \multicolumn{3}{c}{$\times 4$} \\
        \cline{2-11}
        & Method & PSNR & SSIM & LR-PSNR & PSNR & SSIM & LR-PSNR & PSNR & SSIM & LR-PSNR\\
        \hline\hline
        \multirow{4}{4em}{DIV2K}& Bicubic & 31.06 & 0.8937 & 40.38 & 28.26 & 0.8138 & 39.45 & 26.7 & 0.753 & 38.68\\
        & Meta-SR \cite{hu2019meta} & \bf{34.64} & \bf{0.938} & \bf{57.13} & 30.91 & 0.8736 & \bf{54.93} & 28.89 & 0.8173 & \bf{53.19} \\
        & LIIF \cite{chen2021learning} & 34.46 & 0.9367 & 53.63 & 30.81 & 0.8724 & 52.23 & 28.88 & 0.8163 & 50.95\\ 
        & DIINN (ours) & 34.63 & \bf{0.938} & 55.96 & \bf{30.93} & \bf{0.8741} & 53.65 & \bf{28.98} & \bf{0.8193} & 52 \\  
        \hline
        \multirow{4}{4em}{B100} & Bicubic & 28.27 & 0.8316 & 38.54 & 25.88 & 0.7171 & 38.19 & 24.64 & 0.6411 & 37.93\\
        & Meta-SR \cite{hu2019meta} & 29.23 & 0.8651 & 41.5 & 27.51 & 0.7823 & 48.6 & 25.94 & 0.701 & 47.36\\
        & LIIF \cite{chen2021learning} & 30.66 & 0.8891 & 52.27 & 27.68 & 0.7862 & 51.67 & 26.17 & 0.7097 & 50.61\\ 
        & DIINN (ours) & \bf{30.69} & \bf{0.8896} & \bf{52.62} & \bf{27.73} & \bf{0.7873} & \bf{52.36} & \bf{26.22} & \bf{0.7119} & \bf{50.82}\\ 
            
        \hline
        \multirow{4}{4em}{Set5} & Bicubic & 31.81 & 0.9097 & 40.77 & 28.63 & 0.8385 & 38.79 & 26.7 & 0.7739 & 37.29\\
        & Meta-SR \cite{hu2019meta} & \bf{35.68} & \bf{0.9439} & \bf{57.21} & 31.6 & 0.8878 & 48.82 & 29.95 & 0.8605 & \bf{53.79}\\
        & LIIF \cite{chen2021learning} & 35.5 & 0.9427 & 54.06 & 32.15 & 0.9008 & 52.61 & 29.92 & 0.8601 & 50.52\\ 
        & DIINN (ours) & 35.67 & 0.9438 & 56.38 & \bf{32.26} & \bf{0.902} & \bf{54.2} & \bf{30.06} & \bf{0.8631} & 51.89\\
        \hline
        \multirow{4}{4em}{Set14}& Bicubic & 28.33 & 0.8437 & 38.25 & 25.74 & 0.7413 & 37.18 & 24.24 & 0.6648 & 36.57\\
        & Meta-SR \cite{hu2019meta} & 30.9 & 0.8897 & 52.93 & 27.65 & 0.8005 & 46.51 & 26.25 & 0.7383 & \bf{50.61}\\
        & LIIF \cite{chen2021learning} & 31.15 & 0.8919 & 51.57 & 28.04 & 0.8083 & 50.43 & 26.34 & 0.7407 & 49.22\\ 
        & DIINN (ours) & \bf{31.29} & \bf{0.8937} & \bf{53.92} & \bf{28.14} & \bf{0.8101} & \bf{51.53} & \bf{26.43} & \bf{0.7437} & 50.33\\  
        \hline
        \multirow{4}{4em}{Urban100} & Bicubic & 25.44 & 0.8284 & 35.54 & 23.01 & 0.7151 & 35.02 & 21.69 & 0.6334 & 34.64\\
        & Meta-SR \cite{hu2019meta} & 29.32 & 0.907 & 47.53 & 25.53 & 0.8152 & 42.79 & 24.12 & 0.7507 & 48.35\\
        & LIIF \cite{chen2021learning} & 30.02 & 0.9147 & 48.92 & 26.29 & 0.8296 & 48.34 & 24.29 & 0.7555 & 47.33\\ 
        & DIINN (ours) & \bf{30.29} & \bf{0.9176} & \bf{51.2} & \bf{26.46} & \bf{0.8337} & \bf{49.67} & \bf{24.49} & \bf{0.7624} & \bf{48.75}\\ 
    \end{tabular}
\end{adjustbox}
\caption{A comparison against state-of-the-art SISR methods that allow for
arbitrary scale upsampling at trained scales.} 
\label{tab:benchmark_results_trained_scales}
\end{table*}

\begin{table*}
\centering
\begin{adjustbox}{scale=.9}
    \begin{tabular}{c|c c c | c c c}
        Scale & \multicolumn{3}{c|}{$\times 2.5$} & \multicolumn{3}{c}{$\times 3.5$} \\
        \cline{1-7}
        Method & PSNR & SSIM & LR-PSNR & PSNR & SSIM & LR-PSNR\\
        \hline\hline
        Bicubic & 29.41 & 0.8514 & 39.89 & 27.39 & 0.781 & 39.02\\
        Meta-SR \cite{hu2019meta} & 31.36 & 0.8898 & 46.1 & 29.21 & 0.8305 & 47.18\\
        LIIF \cite{chen2021learning} & 32.29 & 0.9036 & 52.93 & 29.73 & 0.8427 & 51.62\\ 
        DIINN (ours) & \bf{32.34} & \bf{0.9046} & \bf{53.43} & \bf{29.82} & \bf{0.845} & \bf{52.39}\\  
    \end{tabular}
\end{adjustbox}
\caption{A comparison against state-of-the-art SISR methods that allow for
arbitrary scale upsampling on the DIV2K dataset at ``inter''-scales.} 
\label{tab:benchmark_results_inter_scales}
\end{table*}

Compared to the closest competing framework, LIIF \cite{chen2021learning}, our
method uses a more expressive neural network for the decoder, which leads to
better performance. Unlike LIIF where the content features ($z$) and positional
features ($p$) are concatenated and fed to a single-branch decoder, we decouple
these features and use two separate branches while allowing for interactions
between the two branches. We show in Section~\ref{sec:experiments} that our
method is both \textit{better} and \textit{faster}.

\subsection{Architecture Details}
Note that while we present our method in a point-wise manner, in practice we
implement the model to process the whole image. We use RDN
\cite{zhang2018residual}, without their upsampling module, as the encoder. Our
decoder consists of two 4-layer MLPs (each with 256 hidden units) as described
in Section~\ref{subsec:method-decoder}. We implement the MLPs as convolutional
layers with 256 kernels of size 1. As discussed in
Section~\ref{subsec:method-encoder}, after feeding the encoder with the LR
image, we obtain a deep feature map of the same spatial size with 64 channels.
Then, we perform the nearest interpolation thus effectively increasing the
spatial size of the deep feature map to be the same as the target resolution.

The right-hand side of \eqref{eq:mpn1} takes this upsampled deep feature map
(denoted $z$) as input. Then, we construct a 2D grid of \textit{global}
coordinates and compute the respective \textit{local} coordinates, which results
in a tensor of the same spatial size as the target resolution with 2 channels
for $x$ and $y$. We additionally attach to it $\frac{1}{upscale\_ratio}$ and
denote the concatenation as $p$. \eqref{eq:mpn2} takes this concatenated tensor
as input. Finally, we pass the output of the synthesis network ($s_{N-1}$)
through a convolutional layer with a kernel size of 1 to produce the predicted
SR image.

\begin{table*}
\centering
\begin{adjustbox}{max width=.95\textwidth}
\begin{tabular}{c|c c c | c c c | c c c | c c c}
  Scale & \multicolumn{3}{c|}{$\times 6$} & \multicolumn{3}{c|}{$\times 8$} & \multicolumn{3}{c|}{$\times 10$} & \multicolumn{3}{c}{$\times 15$}\\
  \cline{1-13}
  Method & PSNR & SSIM & LR-PSNR & PSNR & SSIM & LR-PSNR & PSNR & SSIM & LR-PSNR & PSNR & SSIM & LR-PSNR\\
  \hline\hline
  Bicubic & 24.87 & 0.6761 & 37.75 & 23.75 & 0.6326 & 37.14 & 22.95 & 0.6056 & 36.72 & 21.63 & 0.57 & 36.02\\
  Meta-SR \cite{hu2019meta} & 26.53 & 0.7294 & 49.51 & 25.13 & 0.6704 & 46.64 & 24.18 & 0.633 & 45.16 & 22.63 & 0.5819 & 43.27\\
  LIIF \cite{chen2021learning} & 26.65 & 0.7368 & 49.3 & 25.32 & 0.6865 & 48.3 & 24.38 & 0.6527 & 47.55 & 22.82 & 0.6038 & 46.49\\ 
  DIINN (ours) & \bf{26.74} & \bf{0.7404} & \bf{50.18} & \bf{25.41} & \bf{0.6898} & \bf{49.17} & \bf{24.46} & \bf{0.6557} & \bf{48.48} & \bf{22.89} & \bf{0.6063} & \bf{47.55}\\  
\end{tabular}
\end{adjustbox}
\caption{A comparison against state-of-the-art SISR methods that allow for
arbitrary scale upsampling on the DIV2K dataset at ``outer''-scales.} 
\label{tab:benchmark_results_outer_scales}
\end{table*}

\begin{table*}
\centering
\begin{adjustbox}{scale=.9}
\begin{tabular}{c|c c c}
  Output size & $(128\times128)$ & $(256\times256)$ & $(512\times512)$\\
  \hline
  Method & \multicolumn{3}{c}{Runtime (ms)}\\
  \hline\hline
  Bicubic & 0.11 & 0.15 & 0.49\\
  Meta-SR \cite{hu2019meta} & 50.27 & 69.27 & 162.22\\
  LIIF \cite{chen2021learning} & 73.36 & 185.42 & 693.27\\
  DIINN (ours) & 56.6 & 95.51 & 243.59
\end{tabular}
\end{adjustbox}
\caption{A runtime comparison in milliseconds. The input is a single RGB image
of size $(48\times48)$. We report the average runtime of the forward pass for
each method and SR size over 100 runs on a single NVIDIA Quadro RTX 3000.
Bicubic is for reference only.} 
\label{tab:runtime}
\end{table*}

%%%%%%%%%%%%%%%%%%%%%%%%%%%%%%%%%%%%%%%%%%%%%%%%%%%%%%%%%%%%%%%%%%%%%%%%%%%%%%%%
\section{Experiments} 
\label{sec:experiments}
In this section, we present an experimental evaluation of DIINN for SISR. We
give an overview of the datasets and metrics used in
Section~\ref{subsec:datasets_and_metrics}, along with the training details in
Section~\ref{subsec:training_details}. In
Section~\ref{subsec:benchmark_results}, we highlight our benchmark results on
popular datasets and provide a comparison with state-of-the-art methods. We
discuss and illustrate our qualitative results on various image scales in
Section~\ref{subsec:qualitative_results}. Finally, we present an ablation study
in Section~\ref{subsec:ablation_study}.

\subsection{Datasets and Metrics}
\label{subsec:datasets_and_metrics}
The DIV2K dataset \cite{agustsson2017ntire}, released for the NTIRE 2017
challenge on SISR \cite{timofte2017ntire}, consists of 1000 HR images each of
which has a height or width equal to 2040. Our model is trained using a split of
800 HR images. We do not retrain and we preserve all hyperparameters when
testing the model on the following four standard benchmark datasets: DIV2K
validation set (100 HR images), Set5, Set14, BSDS100 \cite{martin2001database},
and Urban100 \cite{huang2015single}. The results are evaluated using the peak
signal-to-noise ratio (PSNR) \cite{irani1993motion} and the structural
similarity index measure (SSIM) \cite{wang2004ssim}. Additionally, we evaluate
the LR consistency by measuring the LR-PSNR. LR-PSNR is computed as the PSNR
between the downsampled SR image and the downsampled ground-truth HR image with
the same bicubic kernel.

\subsection{Training Details}
\label{subsec:training_details}
Our training procedure is similar to \cite{hu2019meta,chen2021learning}. For
each scale factor $s \in \{2,3,4\}$ and HR image in the minibatch, we randomly
cropped a $48s\times 48s$ patch and downsample it using bicubic interpolation
via the \textit{resize} function available in \textit{Torchvision}
\cite{paszke2019pytorch}. We randomly applied horizontal, vertical, and/or
diagonal flips, each with a probability of 0.5. A minibatch of 4 HR images was
used, which resulted in 12 pairs of LR and HR images across the three scales.
We trained our model for 1000 epochs, where each epoch is a full pass through
800 HR images in the DIV2K training set, using the Adam \cite{kingma2015adam}
optimizer with the default hyperparameters provided by \textit{PyTorch}
\cite{paszke2019pytorch}. The learning rate was initialized at $10^{-4}$ and
halved every 200 epochs. To ensure a fair comparison, we followed previous works
\cite{hu2019meta,chen2021learning} and used the L1 loss to train our network.

\subsection{Benchmark Results} 
\label{subsec:benchmark_results}
We compared DIINN against other methods that allow for arbitrary scale
upsampling, namely, Meta-SR \cite{hu2019meta} and LIIF \cite{chen2021learning}.
We retrained Meta-SR and LIIF under the same settings. In
Tables~\ref{tab:benchmark_results_trained_scales}-\ref{tab:benchmark_results_outer_scales},
we show the results on the upsampling scales that the models were trained with
(\eg, $\times2,\times3,$ and $\times4$),  ``inter''-scales (\eg, $\times2.5$ and
$\times3.5$), and the ``outer''-scales (\eg, $\times6, \times8, \times10,$ and
$\times15$). The HR images were downsampled by each scale to be used as inputs
and the SR images are of the same size as the HR images.  

For the scales that the models were trained with
(Table~\ref{tab:benchmark_results_trained_scales}), we observe that in terms of
PSNR and SSIM, DIINN only underperforms Meta-SR in two settings ($\times2$ scale
for DIV2K and Set5) while it outperforms both Meta-SR and LIIF in all other
settings. For both ``inter''-scales
(Table~\ref{tab:benchmark_results_inter_scales}) and ``outer''-scales
(Table~\ref{tab:benchmark_results_outer_scales}), DIINN consistently beats both
Meta-SR and LIIF across all settings. In addition, we report the inference time
of all the methods with the same input and output sizes in
Table~\ref{tab:runtime}. The results confirm that our framework is competitive
with the state of the art in both \textit{performance} and
\textit{generalizability}.

\subsection{Qualitative Results}
\label{subsec:qualitative_results}
\begin{figure*}
\includegraphics[width=.95\textwidth]{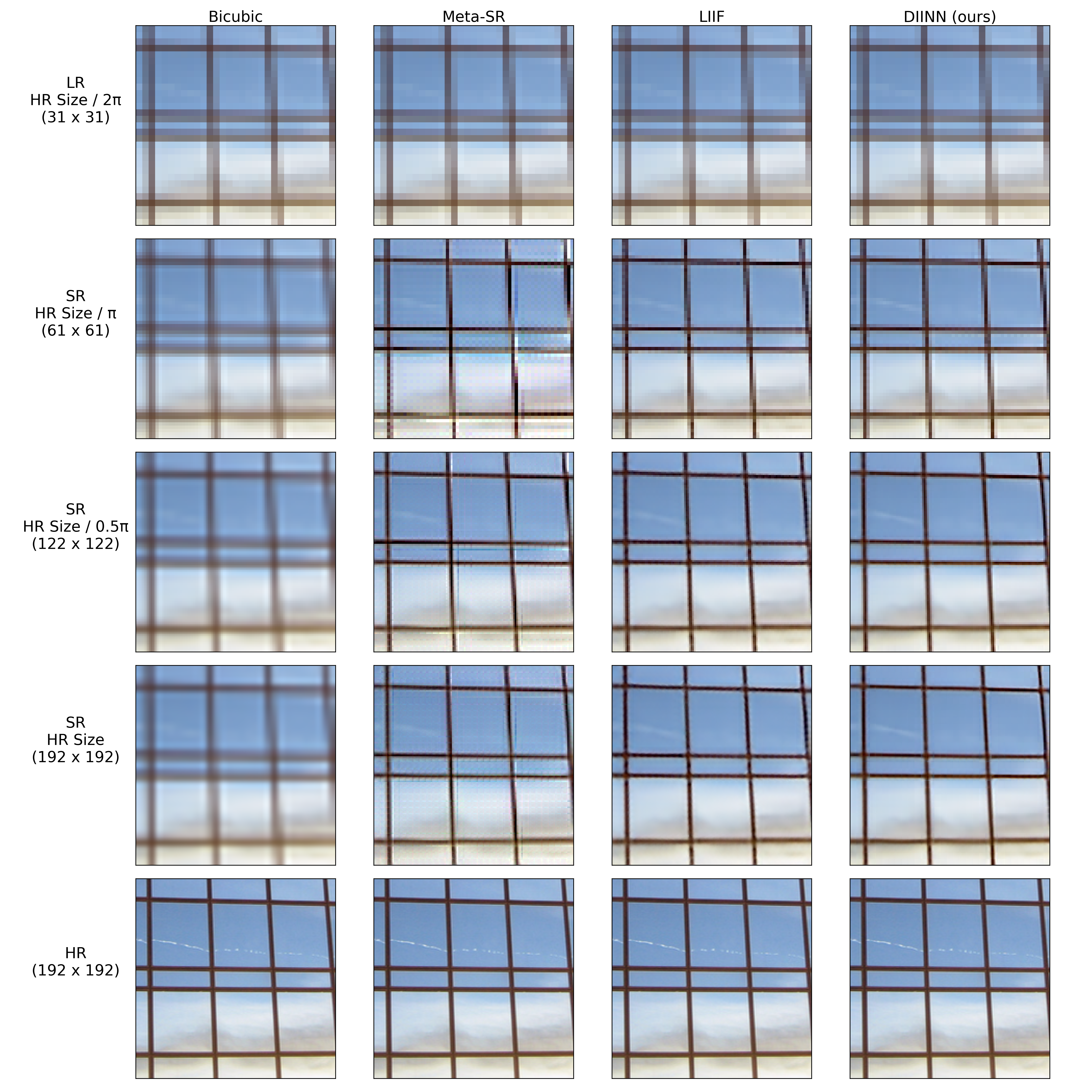}
\caption{Qualitative results on an image patch from the Urban100 dataset.}
\label{fig:qualitative_results}
\end{figure*}

In Figure~\ref{fig:qualitative_results}, we show a qualitative comparison
between bicubic interpolation (first column), Meta-SR (second column), LIIF
(third column), and DIINN (last column). SR predictions are obtained across all
methods from the same LR input (downsampled via bicubic interpolation by a
factor of $2\pi\approx6.28$) at increasing scale factors. We denote the size of
the images for each row and the method for each column. As expected, bicubic
interpolation only smooths the LR image and is unable to recover sharp details.
SR images predicted by Meta-SR have artifacts, especially around the boundary
locations corresponding to the edges of LR pixels. We observe significantly
better results from LIIF and DIINN, which shows the benefits of implicit
decoders for SISR. 

Compared to LIIF, DIINN produces sharper edges and details. Note that if we zoom
in on the results of our method, we can see similar artifacts to those of
Meta-SR, but much less distinguishable. LIIF overcomes these types of artifacts
by averaging their predictions over a small window, which is termed ``local
ensemble''. However, as we observe in the third column in
Figure~\ref{fig:qualitative_results}, the ``local ensemble'' introduces blunter
edges at a higher computational cost. We also note that the aforementioned
artifacts do not appear visually with DIINN when the LR input is of higher
quality (\eg, downsampled HR image at a factor of $4$ or less). 

\begin{table*}[!t]
    \centering
    \begin{adjustbox}{scale=.9}
        \begin{tabular}{c c c|c c c |c c c | c c c}
            \multicolumn{3}{c|}{Scale} & \multicolumn{3}{c|}{$\times 3.14$} & \multicolumn{3}{c|}{$\times 4$} & \multicolumn{3}{c}{$\times 8$} \\
            \hline
            Model & MI & IP & PSNR & SSIM & LR-PSNR & PSNR & SSIM & LR-PSNR & PSNR & SSIM & LR-PSNR\\
            \hline\hline
            (a) & $[m]$ & Yes & 25.67 & 0.8105 & 47.5 & 24.07 & 0.7465 & 47.25 & 20.68 & 0.5643 & 45.75\\
            (b) & $[m]$ & No & 25.78 & 0.814 & 47.93 & 24.17 & 0.7506 & 48.02 & 20.74 & 0.5673 & 46.42\\
            \hline
            (c) & $[m\;z]$ & Yes & 25.89 & 0.8168 & 48.73 & 24.26 & 0.7538 & 48.26 & 20.77 & 0.5704 & 46.26\\
            (d) & $[m\;z]$ & No & 26.07 & 0.8218 & 49.11 & 24.42 & 0.7598 & 48.88 & 20.87 & 0.5762 & 47.05\\
            \hline
            (e) & $[s\;z]$ & Yes & 25.93 & 0.8179 & 48.58 & 24.3 & 0.7555 & 48.12 & 20.8 & 0.5725 & 45.95\\
            (f) & $[s\;z]$ & No & 26.14 & 0.8233 & 49.4 & 24.49 & 0.7624 & 48.75 & 20.91 & 0.5798 & 46.23\\
        \end{tabular}
    \end{adjustbox}
    \caption{A quantitative ablation study on the Urban100 dataset. ``MI'' stands for
        the modulation inputs (\ie, the inputs to the subsequent modulation layers) and
        ``IP'' stands for the initialization of the positional features ($p$).} 
    \label{tab:ablation_0}
\end{table*}

\subsection{Ablation Study} 
\label{subsec:ablation_study}
To more closely examine the design choices of our architecture, we performed an
ablation study on variations of the dual implicit decoder. In
Table~\ref{tab:ablation_0}, we summarize the quantitative results of six
variations on the Urban100 dataset. Note that model (f) is our final model,
which we used to report the results in the previous tables.

\subsubsection{Modulation Inputs} 
\label{subsubsec:modulation_inputs}
First, we adjust the inputs to the modulation network in following three ways.
\begin{enumerate}
  \item For models (a) and (b), each subsequent modulation layer takes only the
  output of its previous modulation layer as input and the residual connections
  to the content features ($z$) are removed. The following equation is used in
  place of \eqref{eq:mpn3}:
  \begin{equation}
    m_i = ReLU(w_i m_{i-1} + b_i).
  \end{equation}
  \item For models (c) and (d), each subsequent modulation layer takes a
  concatenation of the output of its previous modulation layer and the content
  features as input, as proposed in \cite{mehta2021modulated}. The following
  equation is used in place of \eqref{eq:mpn3}:
  \begin{equation} \label{eq:model_cd}
    m_i = ReLU(w_i [m_{i-1} \; z] + b_i).
  \end{equation}
  \item For models (e) and (f), each subsequent modulation layer takes a
  concatenation of the output of the previous synthesis layer and the content
  features as input, as expressed in \eqref{eq:mpn3}.
\end{enumerate}
Comparing models (a) and (c), or models (b) and (d), we observe that the
addition of the skip connections to the content features ($z$) increases the
SISR performance similar to the observation in \cite{mehta2021modulated} on
image reconstruction. This performance boost comes with additional parameters in
the modulation networks. Nonetheless, comparing models (c) and (e), or models
(d) and (f), we find that by simply using the output of the previous synthesis
layer instead of the previous modulation layer, we can increase the PSNR and
SSIM performance at no additional cost. We note that the consistency of the SR
images with respect to the LR inputs, measured by the LR-PSNR, is better with
models (c) and (d) compared to models (e) and (f) at larger scale factors. 

\subsubsection{Positional Features} 
\label{subsubsec:positional_features}
Next, we explore a way to connect the positional features with the neighborhood
around the corresponding LR pixels introduced by the feature unfolding operation
(Section~\ref{subsec:method-encoder}). We attempt to do this by adding an
initializing dense layer with a sine activation at the beginning of the decoder
(\ie, before we perform \eqref{eq:mpn1}). The following equations are used to
transform both the positional and the content features:
\begin{align}
  p &\gets \sin(w_i p + b_i), \label{eq:init_q1}\\
  z &\gets p \odot z. \label{eq:init_q2}
\end{align}
Intuitively, we want this initializing layer to decide the weights for each LR
pixel in the corresponding neighborhood, which could be seen as a learnable
distance weighting function. As shown in Table~\ref{tab:ablation_0}, we observe
no improvement across the three variations
(Section~\ref{subsubsec:modulation_inputs}) despite allowing for more
flexibility with additional parameters in our model.

%%%%%%%%%%%%%%%%%%%%%%%%%%%%%%%%%%%%%%%%%%%%%%%%%%%%%%%%%%%%%%%%%%%%%%%%%%%%%%%%
\section{Conclusion} 
\label{sec:conclusion}
In this paper, we leveraged the learning of implicit representations to develop
a dual interactive implicit network for SISR. Our approach allows for arbitrary
scale factors without the need to train multiple models. Through extensive
experimental evaluations across many settings and datasets, we empirically
showed that DIINN achieves state-of-the-art results. Exploring ways to transform
our architecture to a larger receptive field along with improved interactions
between the encoder and the decoder will be left for future work.  We will also
consider adapting our method to learn a space of predictions to address the
ill-posed nature of SISR.

\section*{Acknowledgments}                                                                                                                   
The authors acknowledge the Texas Advanced Computing Center (TACC) at the
University of Texas at Austin for providing software, computational, and storage
resources that have contributed to the research results reported within this
paper. 

{\small
\bibliographystyle{ieee_fullname}
\bibliography{sisr}

\begin{thebibliography}{10}\itemsep=-1pt

\bibitem{agustsson2017ntire}
Eirikur Agustsson and Radu Timofte.
\newblock Ntire 2017 challenge on single image super-resolution: Dataset and
  study.
\newblock In {\em Proceedings of the IEEE/CVF Conference on Computer Vision and
  Pattern Recognition Workshops}, pages 126--135, 2017.

\bibitem{atzmon2020sal}
Matan Atzmon and Yaron Lipman.
\newblock Sal: Sign agnostic learning of shapes from raw data.
\newblock In {\em Proceedings of the IEEE/CVF Conference on Computer Vision and
  Pattern Recognition}, pages 2565--2574, 2020.

\bibitem{bai2020survey}
K Bai, X Liao, Q Zhang, X Jia, and S Liu.
\newblock Survey of learning based single image super-resolution reconstruction
  technology.
\newblock {\em Pattern Recognition and Image Analysis}, 30(4):567--577, 2020.

\bibitem{bemana2020x}
Mojtaba Bemana, Karol Myszkowski, Hans-Peter Seidel, and Tobias Ritschel.
\newblock X-fields: Implicit neural view-, light-and time-image interpolation.
\newblock {\em ACM Transactions on Graphics}, 39(6):1--15, 2020.

\bibitem{chabra2020deep}
Rohan Chabra, Jan~E Lenssen, Eddy Ilg, Tanner Schmidt, Julian Straub, Steven
  Lovegrove, and Richard Newcombe.
\newblock Deep local shapes: Learning local sdf priors for detailed 3d
  reconstruction.
\newblock In {\em Proceedings of the European Conference on Computer Vision},
  pages 608--625. Springer, 2020.

\bibitem{chen2021learning}
Yinbo Chen, Sifei Liu, and Xiaolong Wang.
\newblock Learning continuous image representation with local implicit image
  function.
\newblock In {\em Proceedings of the IEEE/CVF Conference on Computer Vision and
  Pattern Recognition}, pages 8628--8638, 2021.

\bibitem{chen2019learning}
Zhiqin Chen and Hao Zhang.
\newblock Learning implicit fields for generative shape modeling.
\newblock In {\em Proceedings of the IEEE/CVF Conference on Computer Vision and
  Pattern Recognition}, pages 5939--5948, 2019.

\bibitem{chibane2020implicit}
Julian Chibane, Thiemo Alldieck, and Gerard Pons-Moll.
\newblock Implicit functions in feature space for 3d shape reconstruction and
  completion.
\newblock In {\em Proceedings of the IEEE/CVF Conference on Computer Vision and
  Pattern Recognition}, pages 6970--6981, 2020.

\bibitem{chibane2020ndf}
Julian Chibane, Aymen Mir, and Gerard Pons-Moll.
\newblock Neural unsigned distance fields for implicit function learning.
\newblock In {\em Proceedings of the Advances in Neural Information Processing
  Systems}, pages 21638--21652, 2020.

\bibitem{diinn2023}
\url{https://github.com/robotic-vision-lab/Dual-Interactive-Implicit-Neural-Network}.

\bibitem{dong2015image}
Chao Dong, Chen~Change Loy, Kaiming He, and Xiaoou Tang.
\newblock Image super-resolution using deep convolutional networks.
\newblock {\em IEEE Transactions on Pattern Analysis and Machine Intelligence},
  38(2):295--307, 2015.

\bibitem{dong2016accelerating}
Chao Dong, Chen~Change Loy, and Xiaoou Tang.
\newblock Accelerating the super-resolution convolutional neural network.
\newblock In {\em Proceedings of the European Conference on Computer Vision},
  pages 391--407. Springer, 2016.

\bibitem{goodfellow2014generative}
Ian Goodfellow, Jean Pouget-Abadie, Mehdi Mirza, Bing Xu, David Warde-Farley,
  Sherjil Ozair, Aaron Courville, and Yoshua Bengio.
\newblock Generative adversarial nets.
\newblock {\em Proceedings of the Advances in Neural Information Processing
  Systems}, 27, 2014.

\bibitem{haris2018deep}
Muhammad Haris, Gregory Shakhnarovich, and Norimichi Ukita.
\newblock Deep back-projection networks for super-resolution.
\newblock In {\em Proceedings of the IEEE/CVF Conference on Computer Vision and
  Pattern Recognition}, pages 1664--1673, 2018.

\bibitem{hu2019meta}
Xuecai Hu, Haoyuan Mu, Xiangyu Zhang, Zilei Wang, Tieniu Tan, and Jian Sun.
\newblock Meta-sr: A magnification-arbitrary network for super-resolution.
\newblock In {\em Proceedings of the IEEE/CVF Conference on Computer Vision and
  Pattern Recognition}, pages 1575--1584, 2019.

\bibitem{huang2015single}
Jia-Bin Huang, Abhishek Singh, and Narendra Ahuja.
\newblock Single image super-resolution from transformed self-exemplars.
\newblock In {\em Proceedings of the IEEE/CVF Conference on Computer Vision and
  Pattern Recognition}, pages 5197--5206, 2015.

\bibitem{irani1993motion}
Michal Irani and Shmuel Peleg.
\newblock Motion analysis for image enhancement: Resolution, occlusion, and
  transparency.
\newblock {\em Journal of Visual Communication and Image Representation},
  4(4):324--335, 1993.

\bibitem{jiang2020local}
Chiyu Jiang, Avneesh Sud, Ameesh Makadia, Jingwei Huang, Matthias Nie{\ss}ner,
  Thomas Funkhouser, et~al.
\newblock Local implicit grid representations for 3d scenes.
\newblock In {\em Proceedings of the IEEE/CVF Conference on Computer Vision and
  Pattern Recognition}, pages 6001--6010, 2020.

\bibitem{johnson2016perceptual}
Justin Johnson, Alexandre Alahi, and Li Fei-Fei.
\newblock Perceptual losses for real-time style transfer and super-resolution.
\newblock In {\em Proceedings of the European Conference on Computer Vision},
  pages 694--711. Springer, 2016.

\bibitem{kim2016accurate}
Jiwon Kim, Jung~Kwon Lee, and Kyoung~Mu Lee.
\newblock Accurate image super-resolution using very deep convolutional
  networks.
\newblock In {\em Proceedings of the IEEE/CVF Conference on Computer Vision and
  Pattern Recognition}, pages 1646--1654, 2016.

\bibitem{kim2016deeply}
Jiwon Kim, Jung~Kwon Lee, and Kyoung~Mu Lee.
\newblock Deeply-recursive convolutional network for image super-resolution.
\newblock In {\em Proceedings of the IEEE/CVF Conference on Computer Vision and
  Pattern Recognition}, pages 1637--1645, 2016.

\bibitem{kingma2015adam}
Diederik~P. Kingma and Jimmy Ba.
\newblock Adam: A method for stochastic optimization.
\newblock In {\em Proceedings of the International Conference on Learning
  Representations}, 2015.

\bibitem{lim2017enhanced}
Bee Lim, Sanghyun Son, Heewon Kim, Seungjun Nah, and Kyoung Mu~Lee.
\newblock Enhanced deep residual networks for single image super-resolution.
\newblock In {\em Proceedings of the IEEE/CVF Conference on Computer Vision and
  Pattern Recognition}, pages 136--144, 2017.

\bibitem{ma2022recovering}
Cheng Ma, Peiqi Yu, Jiwen Lu, and Jie Zhou.
\newblock Recovering realistic details for magnification-arbitrary image
  super-resolution.
\newblock {\em IEEE Transactions on Image Processing}, 2022.

\bibitem{martin2001database}
David Martin, Charless Fowlkes, Doron Tal, and Jitendra Malik.
\newblock A database of human segmented natural images and its application to
  evaluating segmentation algorithms and measuring ecological statistics.
\newblock In {\em Proceedings of the IEEE/CVF International Conference on
  Computer Vision}, volume~2, pages 416--423, 2001.

\bibitem{mehta2021modulated}
Ishit Mehta, Micha{\"e}l Gharbi, Connelly Barnes, Eli Shechtman, Ravi
  Ramamoorthi, and Manmohan Chandraker.
\newblock Modulated periodic activations for generalizable local functional
  representations.
\newblock In {\em Proceedings of the IEEE/CVF International Conference on
  Computer Vision}, pages 14214--14223, 2021.

\bibitem{mescheder2019occupancy}
Lars Mescheder, Michael Oechsle, Michael Niemeyer, Sebastian Nowozin, and
  Andreas Geiger.
\newblock Occupancy networks: Learning 3d reconstruction in function space.
\newblock In {\em Proceedings of the IEEE/CVF Conference on Computer Vision and
  Pattern Recognition}, pages 4460--4470, 2019.

\bibitem{park2019deepsdf}
Jeong~Joon Park, Peter Florence, Julian Straub, Richard Newcombe, and Steven
  Lovegrove.
\newblock Deepsdf: Learning continuous signed distance functions for shape
  representation.
\newblock In {\em Proceedings of the IEEE/CVF Conference on Computer Vision and
  Pattern Recognition}, pages 165--174, 2019.

\bibitem{paszke2019pytorch}
Adam Paszke, Sam Gross, Francisco Massa, Adam Lerer, James Bradbury, Gregory
  Chanan, Trevor Killeen, Zeming Lin, Natalia Gimelshein, Luca Antiga, Alban
  Desmaison, Andreas Kopf, Edward Yang, Zachary DeVito, Martin Raison, Alykhan
  Tejani, Sasank Chilamkurthy, Benoit Steiner, Lu Fang, Junjie Bai, and Soumith
  Chintala.
\newblock Pytorch: An imperative style, high-performance deep learning library.
\newblock {\em Proceedings of the Advances in Neural Information Processing
  Systems}, 32:8024--8035, 2019.

\bibitem{peng2020convolutional}
Songyou Peng, Michael Niemeyer, Lars Mescheder, Marc Pollefeys, and Andreas
  Geiger.
\newblock Convolutional occupancy networks.
\newblock In {\em Proceedings of the European Conference on Computer Vision},
  pages 523--540. Springer, 2020.

\bibitem{ramon2021h3d}
Eduard Ramon, Gil Triginer, Janna Escur, Albert Pumarola, Jaime Garcia, Xavier
  Giro-i Nieto, and Francesc Moreno-Noguer.
\newblock H3d-net: Few-shot high-fidelity 3d head reconstruction.
\newblock In {\em Proceedings of the IEEE/CVF International Conference on
  Computer Vision}, pages 5620--5629, 2021.

\bibitem{shi2016real}
Wenzhe Shi, Jose Caballero, Ferenc Husz{\'a}r, Johannes Totz, Andrew~P Aitken,
  Rob Bishop, Daniel Rueckert, and Zehan Wang.
\newblock Real-time single image and video super-resolution using an efficient
  sub-pixel convolutional neural network.
\newblock In {\em Proceedings of the IEEE/CVF Conference on Computer Vision and
  Pattern Recognition}, pages 1874--1883, 2016.

\bibitem{sitzmann2020implicit}
Vincent Sitzmann, Julien Martel, Alexander Bergman, David Lindell, and Gordon
  Wetzstein.
\newblock Implicit neural representations with periodic activation functions.
\newblock {\em Proceedings of the Advances in Neural Information Processing
  Systems}, 33:7462--7473, 2020.

\bibitem{stanley2007compositional}
Kenneth~O Stanley.
\newblock Compositional pattern producing networks: A novel abstraction of
  development.
\newblock {\em Genetic Programming and Evolvable Machines}, 8(2):131--162,
  2007.

\bibitem{tancik2020fourfeat}
Matthew Tancik, Pratul~P. Srinivasan, Ben Mildenhall, Sara Fridovich-Keil,
  Nithin Raghavan, Utkarsh Singhal, Ravi Ramamoorthi, Jonathan~T. Barron, and
  Ren Ng.
\newblock Fourier features let networks learn high frequency functions in low
  dimensional domains.
\newblock {\em Proceedings of the Advances in Neural Information Processing
  Systems}, 2020.

\bibitem{timofte2017ntire}
Radu Timofte, Eirikur Agustsson, Luc Van~Gool, Ming-Hsuan Yang, and Lei Zhang.
\newblock Ntire 2017 challenge on single image super-resolution: Methods and
  results.
\newblock In {\em Proceedings of the IEEE/CVF Conference on Computer Vision and
  Pattern Recognition Workshops}, pages 114--125, 2017.

\bibitem{wang2020brief}
Wei Wang, Yihui Hu, Yanhong Luo, and Tong Zhang.
\newblock Brief survey of single image super-resolution reconstruction based on
  deep learning approaches.
\newblock {\em Sensing and Imaging}, 21(1):1--20, 2020.

\bibitem{wang2004ssim}
Zhou Wang, A.C. Bovik, H.R. Sheikh, and E.P. Simoncelli.
\newblock Image quality assessment: From error visibility to structural
  similarity.
\newblock {\em IEEE Transactions on Image Processing}, 13(4):600--612, 2004.

\bibitem{yang2017image}
Jianchao Yang and Thomas Huang.
\newblock Image super-resolution: Historical overview and future challenges.
\newblock In {\em Super-Resolution Imaging}, pages 1--34. CRC Press, 2017.

\bibitem{yue2016image}
Linwei Yue, Huanfeng Shen, Jie Li, Qiangqiang Yuan, Hongyan Zhang, and Liangpei
  Zhang.
\newblock Image super-resolution: The techniques, applications, and future.
\newblock {\em Signal Processing}, 128:389--408, 2016.

\bibitem{zhang2018residual}
Yulun Zhang, Yapeng Tian, Yu Kong, Bineng Zhong, and Yun Fu.
\newblock Residual dense network for image super-resolution.
\newblock In {\em Proceedings of the IEEE/CVF Conference on Computer Vision and
  Pattern Recognition}, 2018.

\end{thebibliography}
}

\end{document}